\documentclass[conference]{IEEEtran}
\IEEEoverridecommandlockouts
\usepackage{cite}
\usepackage{amsmath,amssymb,amsfonts}
\usepackage{algorithmic}
\usepackage{graphicx}
\usepackage{textcomp}
\usepackage{xcolor}
\def\BibTeX{{\rm B\kern-.05em{\sc i\kern-.025em b}\kern-.08em
    T\kern-.1667em\lower.7ex\hbox{E}\kern-.125emX}}
\begin{document}

\title{On-Sensor Convolutional Neural Networks with Early-Exits}


  \author{
 \IEEEauthorblockN{
    Hazem Hesham Yousef Shalby\IEEEauthorrefmark{1}\textsuperscript{\textsection},
    Arianna De Vecchi\IEEEauthorrefmark{1}\textsuperscript{\textsection},
    Alice Scandelli\IEEEauthorrefmark{1},
    Pietro Bartoli\IEEEauthorrefmark{1},\\
    Diana Trojaniello\IEEEauthorrefmark{2},
    Manuel Roveri\IEEEauthorrefmark{1} and
    Federica Villa\IEEEauthorrefmark{1}}
 \IEEEauthorblockA{\{hazemhesham.shalby, arianna.devecchi, alice.scandelli, pietro.bartoli, manuel.roveri, federica.villa\}@polimi.it \\
 diana.trojaniello@luxottica.com}

 \IEEEauthorblockA{\IEEEauthorrefmark{1} Politecnico di Milano, Milan, Italy \qquad \IEEEauthorrefmark{2} EssilorLuxottica Smart Eyewear Lab, EssilorLuxottica, Milan, Italy}%
 }

\maketitle
\begingroup\renewcommand\thefootnote{\textsection}
\footnotetext{These authors contributed equally to this work}

\begin{abstract}
Tiny Machine Learning (TinyML) is a novel research field aiming at integrating Machine Learning (ML) within embedded devices with limited memory, computation, and energy.
Recently, a new branch of TinyML has emerged, focusing on integrating ML directly into the sensors to further reduce the power consumption of embedded devices.
Interestingly, despite their state-of-the-art performance in many tasks, none of the current solutions in the literature aims to optimize the implementation of Convolutional Neural Networks (CNNs) operating directly into sensors.
In this paper, we introduce for the first time in the literature the optimized design and implementation of Depth-First CNNs operating on the Intelligent Sensor Processing Unit (ISPU) within an Inertial Measurement Unit (IMU) by STMicroelectronics. Our approach partitions the CNN between the ISPU and the microcontroller (MCU) and employs an Early-Exit mechanism to stop the computations on the IMU when enough confidence about the results is achieved, hence significantly reducing power consumption.
When using a NUCLEO-F411RE board, this solution achieved an average current consumption of 4.8 mA, marking an 11\% reduction compared to the regular inference pipeline on the MCU, while having equal accuracy.
\end{abstract}


\section{Introduction}
Tiny Machine Learning (TinyML) is a novel research field that integrates Machine Learning (ML) models and algorithms within embedded devices, constrained by memory, computation, and power consumption \cite{roveri_is_2023, noauthor_home_nodate,banbury_benchmarking_2021,david_tensorflow_2021}. 
Recently, a new direction in the field has emerged, aimed at pushing ML at the extreme edge, i.e. directly within the sensors \cite{immonen_tiny_2022}. This approach is particularly beneficial for TinyML applications, which usually rely on battery-powered devices, as it further reduces power consumption by utilizing sensors equipped with an Intelligent Sensor Processing Unit (ISPU) \cite{noauthor_lsm6dso16is_nodate}, featuring an ultra-low-power programmable core able to process data within the sensor itself.

In the literature, several works employing the ISPU exist \cite{ronco_machine_2022,immonen_tiny_2022,chowdhary_-sensor_2023,ISPU_gait, SIE_2024}.
Among these solutions, only \cite{SIE_2024} exploits Convolutional Neural Networks (CNNs), which are highly effective across diverse ML tasks. However, in \cite{SIE_2024}, the inference time and the memory occupation depend on the input size as the CNNs are implemented in a Width-First manner. As the input size increases, it exceeds the ISPU’s strict limitations (i.e., a maximum clock frequency of $10\text{ MHz}$ and a memory less than $32\text{ kB}$). Consequently, tasks requiring larger inputs may either surpass the memory limits, making them infeasible, or necessitate a reduced inference frequency for execution.


To address this issue, this paper introduces a novel methodology to design CNNs capable of leveraging the power of ISPUs taking into account their strict memory and computational limits. Specifically, we present three key strategies for achieving this goal:
\begin{enumerate}
    \item \textbf{Depth-First convolutions}: The convolutions within the ISPU are computed using an incremental approach (similar to those explored in the works of \cite{alwani_fused-layer_2016}, \cite{weber_brainslug_2018}, and \cite{binas_low-memory_2019}).
    Specifically, the convolution output, typically computed over multiple samples collected within a temporal window, is updated each time a new sample becomes available.
    This approach removes the need to store all the samples within the window, as the effect of each new sample is computed immediately, further reducing the memory requirements.
    
    \item \textbf{Partitioned Computation}: The CNN computation is partitioned between the ISPU and the microcontroller unit (MCU). This allows us to respect the memory limits of the ISPU, reduce the computational load of the MCU, and reduce the data throughput between the sensor and the MCU. For instance, the ISPU can handle the initial (feature extraction) part of the CNN, while the MCU executes the final part of the CNN.
    
    \item \textbf{Early-Exit (EE) Mechanism}: The EE mechanism interrupts the computational pipeline when a decision can be confidently made without completing the entire pipeline or when certain conditions indicate that completing the computation is unnecessary \cite{scardapane_why_2020,gambella_nachos_2024}. In this context, the ISPU can terminate the computational pipeline based on the confidence about the classification computed on its own process, preventing the need to activate the MCU when unnecessary.
    In this paper, the EE mechanism is implemented as a binary classifier that evaluates whether the MCU must be activated or not. For example, in smart glasses, this could mean activating the MCU only if the glasses are classified as worn by the EE module, and interrupting the computational process if classified as not worn.

\end{enumerate}
We emphasize that by applying these three strategies, our methodology is the first in the literature to enable an efficient implementation of CNNs on the ISPU.

The proposed methodology has been validated using the LSM6DSO16IS 6-axis IMU \cite{noauthor_lsm6dso16is_nodate},  achieving an $11\%$ power reduction compared to the traditional inference pipeline on MCU only when using a NUCLEO-F411RE board. Furthermore, our solution has a memory occupation and an inference time which are independent of the input window size.
In contrast, existing solutions in the literature have a linear relationship between memory usage and inference time relative to window size, which risks exceeding the strict memory constraints of the ISPU in the case of a large input window size (e.g. with a frequency of $26\text{ Hz}$ the maximum window size is $6\text{ s}$).






\section{Related works}\label{sec:related_works}



The ISPU is a core by STMicroelectronics embedded in sensors such as \cite{noauthor_lsm6dso16is_nodate}. It can be programmed in C language to perform data processing at the extreme edge i.e., within the sensor itself.
Such core operates at either $5\text{ MHz}$ or $10\text{ MHz}$ clock frequency and has $8\text{ kB}$ RAM for data storage and $32\text{ kB}$ RAM for program memory.
It is important to notice that the embedded algorithms are triggered at a frequency equal to the Output Data Rate (ODR), i.e. when a new sample becomes available. The ISPU gives precedence to the completion of the algorithms rather than to data acquisition. Therefore, if the processing is not completed before the subsequent sample is available, such sample is lost. 

In the literature, few works exist exploiting the ISPU to bring data processing to the extreme edge.

In \cite{ISPU_gait}, the ISPU was used to perform gait analysis, employing traditional signal processing techniques to extract numerous parameters starting from the acceleration signal in a real-time manner. To prevent data loss, the algorithm was divided into multiple steps.


In \cite{ronco_machine_2022}, an IMU is used to extract features to be given to neural networks, implemented on the ISPU. 
This work aims to differentiate 5 classes, starting from 30 features extracted from 32-sample windows of the IMU data. The feature extraction step requires $6.57 \text{ ms}$ to be completed, while the inference time increases with the complexity of the networks. Moreover, the authors report an energy consumption of 90 $\mu$J per inference.

Finally, in \cite{SIE_2024}, CNNs were implemented in the ISPU to perform Human Activity Recognition (HAR), achieving 99\% accuracy in recognizing 6 activities with $930\text{ nWh}$ energy consumption.
However, out of the $8\text{ kB}$ available for data storage, about $4\text{ kB}$ are occupied to store the input window, leaving a smaller space for the weights of the CNN and prohibiting the usage of complex architectures.
Moreover, such CNNs were not completed without exceeding the timing limit given by the ODR leading to data loss between consecutive windows.


Starting from these studies, to the best of our knowledge, no efficient implementation of CNNs was developed specifically for the ISPU. 


\section{Proposed Methodology}\label{methods} \label{sec:proposed_architecture}
Section \ref{sec:formalization} introduces the used notation, Section \ref{sec:proposed_meth} details the proposed methodology, Section \ref{sec:implementation} focuses on the proposed implementation, and finally Section \ref{sec:training} presents training process.

\subsection{Notation}\label{sec:formalization}
In general, any CNN can be formulated as a function $f(x_t, \ldots,x_{t-(T-1)})$ where:
\begin{itemize}
    \item $x_t \in \mathbb{R}^{N_{in}}$ is the sample acquired at time $t$ of size $N_{in}$, 
    \item $T$ is the size of the window on which $f(\cdot)$ is performed.
\end{itemize}
Moreover, the function $f(\cdot)$ can be further described as:
\[f(x_t,\ldots,x_{t-(T-1)}) = h(g(x_t,\ldots,x_{t-(T-1)}))\]
where $g(x_t,\ldots,x_{t-(T-1)})$ represents the feature extraction process, and $h(\cdot)$ represents the final classification function.

\subsection{Proposed Methodology}\label{sec:proposed_meth}
This section introduces a novel methodology that enables the efficient implementation of CNNs on the ISPU, allowing us to fully leverage the ISPU's capabilities to implement the function $f(\cdot)$. Fig. \ref{fig:overview_schema} reports an overview of the proposed methodology. Specifically, three key strategies are applied:

\begin{figure}[bp!]
    \centering
    \includegraphics[width=0.75\linewidth]{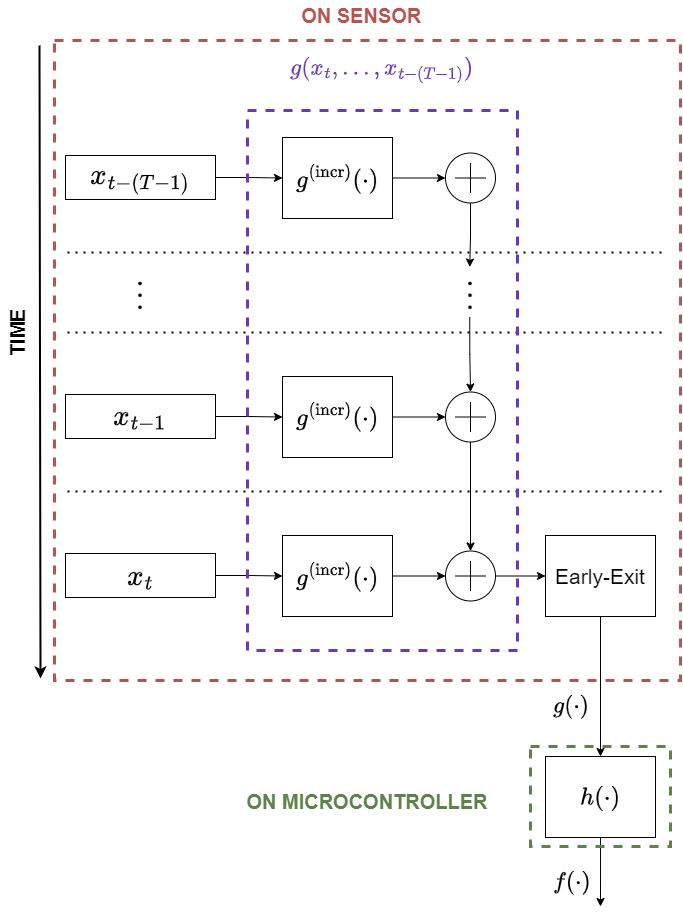}
    \caption{Overview of the proposed approach: \textit{$g(x_t,\ldots,x_{t-(T-1)})$ is the segment of the CNN implemented within the sensor, while $h(\cdot)$ is the one within the MCU. $g^{(incr)}(\cdot)$ is the incremental implementation of $g(\cdot)$. The Early-Exit module triggers the MCU only if needed.}}
    \label{fig:overview_schema}
\end{figure}

\subsubsection{Depth-First convolutions}
Neural networks typically follow a Width-First computation approach, where all operations at one layer must be completed before moving to the next one\cite{weber_brainslug_2018}.
However, this approach has two significant limitations when applied to CNNs on the ISPU: the processing time and memory usage depend on the window size. Consequently, a reduced data sampling rate becomes necessary to prevent sample loss if the ISPU cannot process the data in time. Additionally, a smaller input size is required to fit within the ISPU's memory constraints, even when larger inputs are essential for achieving optimal task performance.

Therefore, we propose an optimization of CNNs computation using an incremental approach, where the output is computed in a Depth-First manner \cite{binas_low-memory_2019, weber_brainslug_2018}. In particular, each time a new sample within the window $T$ becomes available, the output of $g(\cdot)$ is updated. We emphasize that the Depth-First optimization is transparent to the user, since the results of the computations made in this way are equivalent to what is achieved with the Width-First approach.

In this paper, we will focus on tailoring the incremental approach to 1D convolutions.
Nevertheless, it's worth noting that the incremental approach can be extended to all major neural network layers.
Specifically, when $n$ 1D-convolutional filters with a kernel size of $1\times1\times T$ are applied to inputs of size $N_{in}$, they produce an output denoted as
\[Y_t =\begin{bmatrix}
 y^{(1,1)}_t& \ldots & y^{(1,N_{in})}_t\\
\vdots & \ddots & \vdots\\
y^{(n,1)}_t& \ldots & y^{(n,N_{in})}_t\\
\end{bmatrix} \]
where:
\begin{align*}
    y^{(j,-)}_t &= \sum_{i=0}^{T} c_i^{(j)} \times x_{t-i}& \forall j\in[1,n]
\end{align*}
with $c_i^{(j)}$ the weight of the $j$-th filter applied to the sample at time $t-i$.

In the proposed approach, $g^{(incr)}(\cdot)$ is the incremental implementation of $g(\cdot)$. Specifically, $Y_t$ is initialized to $0$, and each time a sample $x_{t-i}$ with $i\in [0, T-1]$ is produced, the following update is performed:
\begin{align*}
    g^{(incr)}(\cdot):\quad y^{(j,-)}_{t-i} &= y^{(j,-)}_{t-i-1} + c_i^{(j)} \times x_{t-i} & \forall j \in [1,n]
\end{align*}

\subsubsection{Partitioned Computation}
$f(\cdot)$ is partitioned between the ISPU and the MCU to maximize the usage of the ISPU capability without exceeding its memory and computational limits.
Specifically, $g(\cdot)$ is processed on the ISPU, while $h(\cdot)$ is processed on the MCU. This division allows us to wake up the MCU only when the output of $g(\cdot)$ has been computed.
In the regular inference pipeline, the MCU is woken up with each new sample $x_{t}$, resulting in inefficient power consumption. To address this, some sensors use a buffer to store multiple samples before activating the MCU, however the buffer has a limited capacity. For instance, the LSM6DSO16IS can store up to 5 samples before generating an interrupt. In contrast, the proposed methodology activates the MCU only when $g(\cdot)$ is computed within the ISPU, thereby reducing the system's energy consumption.

\subsubsection{EE Mechanism}
This module is integrated within the ISPU and decides whether to wake or not the MCU to compute $h(\cdot)$ each time $g(\cdot)$ is computed. 
Specifically, in this paper this module is implemented as a binary classifier (i.e., activate vs not-activate MCU) that evaluates whether specific conditions are met, hence activating the MCU vs interrupting the computation.


\subsection{Implementation}\label{sec:implementation}
The proposed methodology is implemented as follows:
\begin{itemize}
    \item $g(x_t,\ldots,x_{t-(T-1)})$ performs two 1D convolutions with $n=16$ followed by a Max Pooling layer of size $N_{in} = 6$, hence producing an output of dimension 16,
    \item The EE mechanism utilizes a fully connected layer to perform binary classification on the features extracted by $g(\cdot)$. Specifically, this classifier determines whether or not to activate the MCU,
    \item $h(\cdot)$ can be any general task performed on the output of $g(\cdot)$. Here $h(\cdot)$ is left general because the proposed solution can be tailored to any task.
\end{itemize}
The proposed architecture has been chosen after considering the constraints imposed by the ISPU.


\subsection{Training process}\label{sec:training}

The training of the proposed methodology follows a 2-step pipeline. Firstly, to determine the weights of $g(\cdot)$ and $h(\cdot)$, the function $f(\cdot)$ is optimized in an end-to-end manner to minimize a certain loss for the given task. Then, the weights of $g(\cdot)$ are fixed and used to train the EE Module.
Specifically, the EE module is optimized for the task of activating vs not activating the MCU. This requires a dataset of input sequences $\{x_t,\ldots,x_{t-(T-1)}\}$ paired with target labels \{activate, not-activate\}.

\section{Experimental results}\label{sec:evaluation}

This section presents the evaluation results of the proposed methodology, as detailed in Section \ref{sec:implementation}. Specifically, Section \ref{sec:power} discusses power consumption, Section \ref{sec:memory} analyzes memory usage and inference time, and Section \ref{sec:example} provides a real-world scenario evaluation using a simple example implemented on the device.

In the evaluation carried out here, we compared:
\begin{enumerate}
    \item \textbf{Regular pipeline} as \cite{SIE_2024}: the MCU collects the samples and performs the inference,
    \item \textbf{Our pipeline}: the ISPU computes $g(\cdot)$, and the EE decides whether to wake the MCU through an interrupt to read the results of $g(\cdot)$ and compute $h(\cdot)$.
\end{enumerate}
Furthermore, we assessed two sub-cases for each solution: employing the conventional Width-First convolution computation method and employing the Depth-First convolution computation method approach suggested here.


The evaluation was conducted using a NUCLEO-F411RE board with the MCU and the X-NUCLEO-IKS4A1 expansion board, which includes an LSM6DSO16IS, a 6-axis IMU with an accelerometer and gyroscope. The IMU samples data at $26\text{ Hz}$ and performs inferences within 1-second windows.
Additionally, to minimize power consumption, the MCU remained in sleep mode and was activated via interrupts.

\subsection{Power consumption}\label{sec:power}


The current consumption results are as follows.
The MCU in the regular inference pipeline has a current consumption of at least $4.8\text{ mA}$ using the Width-First approach ($5.9\text{ mA}$ using Depth-First computation). Additionally, $0.6\text{ mA}$ are needed by the IMU.
On the other hand, our pipeline results in a decreased current consumption, which is independent of the CNN implementation (Width-First or Depth-First).
In particular, when the EE module activates the MCU, the latter consumes an average of $4.0\text{ mA}$, while the IMU and the ISPU require $0.8\text{ mA}$. This leads to an 11\% reduction with respect to the regular inference pipeline.
Differently, when the EE module interrupts the computation pipeline, the MCU is never woken up and it consumes an average of $3.8\text{ mA}$, while $0.8\text{ mA}$ are used for the IMU and the ISPU. This implies that the current consumption is reduced by 15\% with respect to the regular inference pipeline.


\subsection{Memory occupation and inference time}
\label{sec:memory}

\begin{figure}[bp!]
    \centering
    \includegraphics[width=\linewidth]{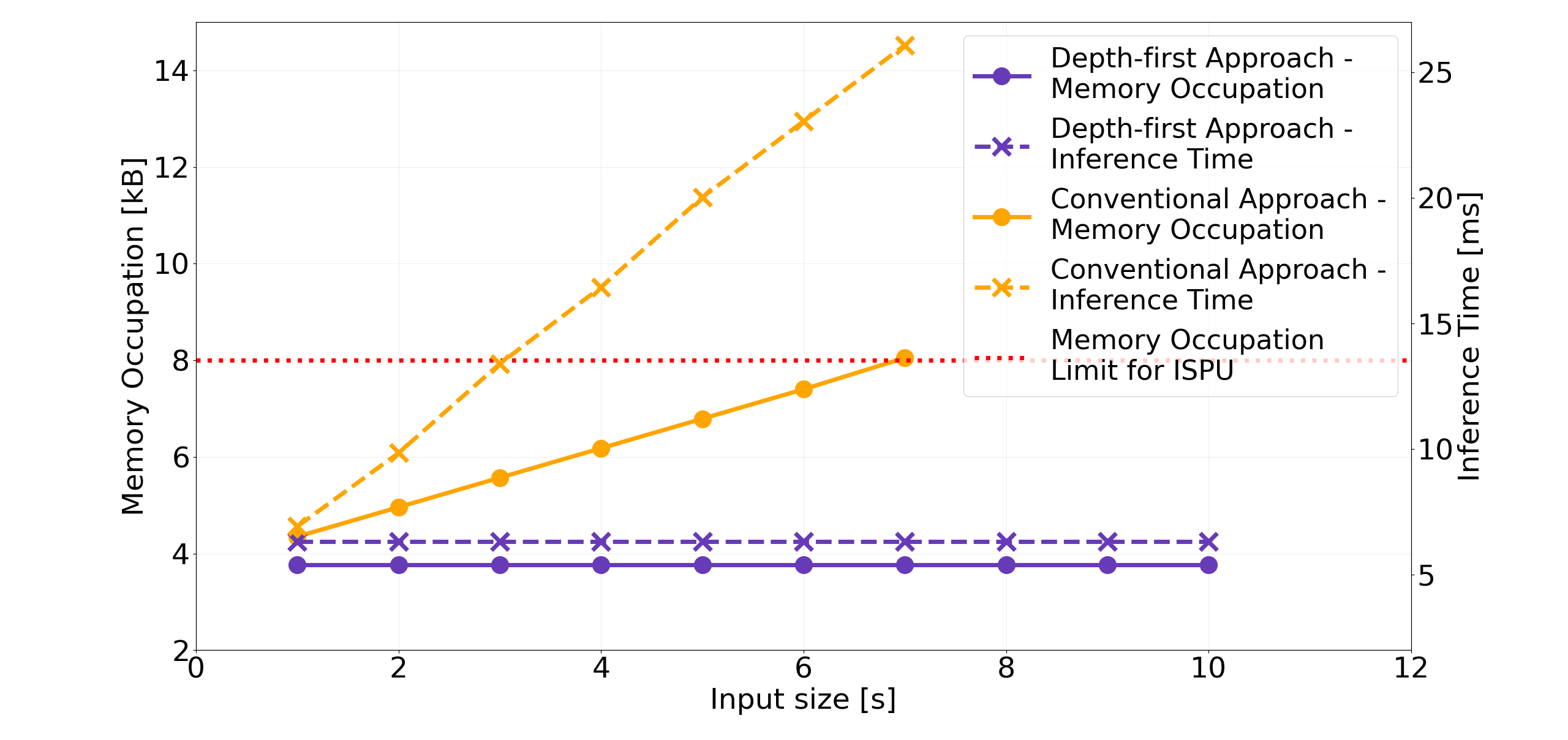}
    \caption{Memory occupation and inference time with respect to input size for the two approaches on ISPU running at 10 MHz and ODR equal to 26 Hz.}
    \label{fig:time_memory_plot}
\end{figure}

The memory occupation and time for the inference evaluation results are reported in Fig. \ref{fig:time_memory_plot}.
Specifically, when using a Width-First approach, both memory consumption and inference time increase together with the size of the input window, leading to a maximum input size of $6\text{ s}$ as larger windows lead to a saturation of the ISPU RAM.
On the other hand, the memory occupation related to our pipeline when using a Depth-First approach remains stable at $3.8\text{ kB}$.

Regarding the inference time, the Depth-First approach requires a maximum of $6.3\text{ ms}$ per inference, independently of the input window size, implying a maximum ODR of $158\text{ Hz}$. On the contrary, when using a Width-First approach with our pipeline, the maximum ODR depends on the input window size. Indeed, with an input window of $6\text{ s}$ (Fig. \ref{fig:time_memory_plot}), the inference time is equal to $23\text{ ms}$, thus the maximum ODR is equal to $43\text{ Hz}$.









\subsection{Proof of Concept example}\label{sec:example}
We evaluated the proposed methodology using a smart eyewear equipped with an LSM6DSO16IS \cite{noauthor_lsm6dso16is_nodate} sensor, which features an ISPU. In this setup, the ISPU is tasked with determining whether the glasses are being worn, and it activates the MCU to elaborate further on the extracted feature only in that case.
The binary classification task of detecting whether glasses are worn is handled by the EE module, implemented as a fully connected layer with two units, and achieved an accuracy ranging from $96\%$ to $99\%$.
Specifically, the EE module is trained on a dataset acquired using smart eyewear embedded with an LSM6DSO16IS having ODR equal to $26\text{ Hz}$ and Full-Scale Ranges (FSRs) equal to $\pm2\text{ g}$ and $\pm250\text{ dps}$. This dataset comprises two distinct classes gathered from 12 participants (9 males, and 3 females, aged $28\pm10$): worn (glasses worn for 3 minutes) and not worn (glasses not worn for 3 minutes).

Taking into account a 1-hour usage of the smart eyewear, if these were worn for 1 hour consecutively, the regular pipeline would lead to an energy consumption of $35\text{ J}$ when powering the system at $1.8\text{ V}$. On the other hand, our pipeline would reduce such value to $31\text{ J}$. Additionally, if the glasses were not worn for 1 hour, utilizing the EE module further cuts the energy usage by over $1\text{ J}$, as the MCU is not woken up.
\renewcommand{\thefootnote}{$1$}
Moreover, taking into account the battery on the smart eyewear used in this study with a capacity of $200\text{ mAh}$, the device would last 37 hours with the regular pipeline, while, when using the proposed methodology, this time would increase by 5 hours, reaching 42 hours of continuous usage when the glasses are worn\footnote{These considerations are done with the assumption that the device is only running the considered example and future development will simulate more complex scenarios and will focus on an ultra-low-power MCU.}.




\section{Conclusions and Future Works} \label{sec:conclusions}
This paper proposes the first implementation of Depth-First CNNs employing an ISPU, significantly reducing power consumption.
Differently from \cite{ronco_machine_2022,SIE_2024}, a new pipeline, optimized for the ISPU, was implemented.
Specifically, the proposed methodology partitions a general CNN between the ISPU (i.e., a core embedded within the sensor) and the MCU. Moreover, using an EE mechanism, the proposed pipeline can interrupt the computation on the ISPU, further reducing the power consumption.
Regarding the results, our implementation led to an $11\%$ reduction in the average current consumption with respect to the regular pipeline. This reduction increases to 15\% when the EE module does not wake up the MCU.
Finally, our implementation allowed the inference time and the memory occupation to be independent of the input window size. This is of utmost importance when programming the ISPU, which must carry out the computation in a time related to the sensor ODR, i.e. before the new samples are available, to avoid data loss.


Future work will focus on more complex tasks, including the development of a function $h(\cdot)$ able to further process the features computed by the ISPU.
Additionally, quantized weights will be integrated into the ISPU to enable support for larger neural network architectures. Finally, the pipeline will be implemented using an ultra-low-power MCU, such as \cite{noauthor_nucleo-U5}, which is expected to significantly enhance energy efficiency and overall system performance compared to the MCU used in this study.


\clearpage
\bibliographystyle{unsrt}
\bibliography{reference}

\begin{thebibliography}{10}

\bibitem{roveri_is_2023}
Manuel Roveri.
\newblock Is {Tiny} {Deep} {Learning} the {New} {Deep} {Learning}?
\newblock In Rajkumar Buyya, Susanna~Munoz Hernandez, Ram Mohan~Rao Kovvur, and T.~Hitendra Sarma, editors, {\em Computational {Intelligence} and {Data} {Analytics}}, pages 23--39, Singapore, 2023. Springer Nature.

\bibitem{noauthor_home_nodate}
Home {\textbar} {tinyML} {Foundation}.

\bibitem{banbury_benchmarking_2021}
Colby~R. Banbury, Vijay~Janapa Reddi, Max Lam, William Fu, Amin Fazel, Jeremy Holleman, Xinyuan Huang, Robert Hurtado, David Kanter, Anton Lokhmotov, David Patterson, Danilo Pau, Jae-sun Seo, Jeff Sieracki, Urmish Thakker, Marian Verhelst, and Poonam Yadav.
\newblock Benchmarking {TinyML} {Systems}: {Challenges} and {Direction}, January 2021.
\newblock arXiv:2003.04821 [cs].

\bibitem{david_tensorflow_2021}
Robert David, Jared Duke, Advait Jain, Vijay~Janapa Reddi, Nat Jeffries, Jian Li, Nick Kreeger, Ian Nappier, Meghna Natraj, Shlomi Regev, Rocky Rhodes, Tiezhen Wang, and Pete Warden.
\newblock {TensorFlow} {Lite} {Micro}: {Embedded} {Machine} {Learning} on {TinyML} {Systems}, March 2021.
\newblock arXiv:2010.08678 [cs].

\bibitem{immonen_tiny_2022}
Riku Immonen and Timo Hämäläinen.
\newblock Tiny {Machine} {Learning} for {Resource}-{Constrained} {Microcontrollers}.
\newblock {\em Journal of Sensors}, 2022:1--11, November 2022.

\bibitem{noauthor_lsm6dso16is_nodate}
{LSM6DSO16IS} - {iNEMO} inertial module: always-on {3D} accelerometer and {3D} gyroscope with {ISPU} - {Intelligent} {Sensor} {Processing} {Unit} - {STMicroelectronics}.

\bibitem{ronco_machine_2022}
Andrea Ronco, Lukas Schulthess, David Zehnder, and Michele Magno.
\newblock Machine {Learning} {In}-{Sensors}: {Computation}-enabled {Intelligent} {Sensors} {For} {Next} {Generation} of {IoT}.
\newblock In {\em 2022 {IEEE} {Sensors}}, pages 01--04, Dallas, TX, USA, October 2022. IEEE.

\bibitem{chowdhary_-sensor_2023}
Mahesh Chowdhary and Swapnil~Sayan Saha.
\newblock On-{Sensor} {Online} {Learning} and {Classification} {Under} 8 {KB} {Memory}.
\newblock In {\em 2023 26th {International} {Conference} on {Information} {Fusion} ({FUSION})}, pages 1--8, June 2023.

\bibitem{ISPU_gait}
Arianna De~Vecchi, Alice Scandelli, Federica Bossi, Benedetta~Caterina Casadei, Marco Boschi, and Federica Villa.
\newblock On-the-edge gait analysis using a smart earable inertial measurement unit.
\newblock In {\em 2024 IEEE Sensors Applications Symposium (SAS)}, pages 1--6, 2024.

\bibitem{SIE_2024}
Arianna De~Vecchi, Alice Scandelli, Federica Bossi, Benedetta~Caterina Casadei, Hazem Hesham~Yousef Shalby, and Federica Villa.
\newblock Efficient human activity recognition: Machine learning at the sensor level.
\newblock In {\em Proceedings of SIE 2024}, 2024.
\newblock in print.

\bibitem{alwani_fused-layer_2016}
Manoj Alwani, Han Chen, Michael Ferdman, and Peter Milder.
\newblock Fused-layer {CNN} accelerators.
\newblock In {\em 2016 49th {Annual} {IEEE}/{ACM} {International} {Symposium} on {Microarchitecture} ({MICRO})}, pages 1--12, Taipei, Taiwan, October 2016. IEEE.

\bibitem{weber_brainslug_2018}
Nicolas Weber, Florian Schmidt, Mathias Niepert, and Felipe Huici.
\newblock {BrainSlug}: {Transparent} {Acceleration} of {Deep} {Learning} {Through} {Depth}-{First} {Parallelism}, April 2018.
\newblock arXiv:1804.08378 [cs].

\bibitem{binas_low-memory_2019}
Jonathan Binas and Yoshua Bengio.
\newblock Low-memory convolutional neural networks through incremental depth-first processing, May 2019.
\newblock arXiv:1804.10727 [cs].

\bibitem{scardapane_why_2020}
Simone Scardapane, Michele Scarpiniti, Enzo Baccarelli, and Aurelio Uncini.
\newblock Why {Should} {We} {Add} {Early} {Exits} to {Neural} {Networks}?
\newblock {\em Cognitive Computation}, 12(5):954--966, September 2020.

\bibitem{gambella_nachos_2024}
Matteo Gambella, Jary Pomponi, Simone Scardapane, and Manuel Roveri.
\newblock {NACHOS}: {Neural} {Architecture} {Search} for {Hardware} {Constrained} {Early} {Exit} {Neural} {Networks}, January 2024.
\newblock arXiv:2401.13330 [cs].

\bibitem{noauthor_nucleo-U5}
{NUCLEO}-{U575ZI} - {STM32} {Nucleo-144} {development} {board} {with} {STM32U575ZIT6Q} {MCU}, {SMPS}, {supports} {Arduino}, {ST} {Zio} {and} {morpho} {connectivity} - {STMicroelectronics}.

\end{thebibliography}

\end{document}